\pdfoutput=1

\documentclass[11pt]{article}

\usepackage[preprint]{acl}

\usepackage{times} 
\usepackage{latexsym}

\usepackage[T1]{fontenc}

\usepackage[utf8]{inputenc}

\usepackage{microtype}

\usepackage{inconsolata}

\usepackage{graphicx}

\usepackage{amsmath}

\usepackage{booktabs}       
\usepackage{amsfonts}       
\usepackage{nicefrac}       
\usepackage{cleveref}       
\usepackage{subcaption}
\usepackage{float}
\usepackage{tcolorbox}
\usepackage{placeins}
\usepackage{relsize}
\usepackage{wrapfig}
\usepackage{hyperref}
\usepackage[size=small]{caption}

\newcounter{textbox}

\newenvironment{textbox}[1][]{%
	\refstepcounter{textbox}%
	\begin{tcolorbox}[colback=lightgray, colframe=black, left=0pt, right=0pt, top=0pt, bottom=0pt, boxrule=0.5pt, arc=2pt, auto outer arc, width=\columnwidth]%
	}{%
	\end{tcolorbox}%
}

\hypersetup{
	pdftitle={COS(M+O)S: Curiosity and RL-Enhanced MCTS for Exploring Story Space via Language Models},
	pdfsubject={cs.CL,cs.AI},
	pdfauthor={Tobias Materzok},
	pdfkeywords={Monte Carlo Tree Search, Large Language Models, Story Generation, Reinforcement Learning, Machine Learning, Preference-Based Fine-Tuning, ORPO, Curiosity},
}

\author{Tobias Materzok \\
  Independent Researcher \\ Darmstadt, Germany \\
  \texttt{t.materzok@theo.chemie.tu-darmstadt.de}
  }

\title{COS(M+O)S: Curiosity and RL-Enhanced MCTS \\ for Exploring Story Space via Language Models}

\begin{document}
\maketitle
\begin{abstract}
\noindent We present COS(M+O)S, a System 2-inspired framework for open-ended plot development that systematically explores the vast space of possible story expansions, enabling a 3B-parameter language model to approach the plot quality of a 70B model on select short-story tasks.
The method accomplishes this by combining Monte Carlo Tree Search (MCTS), guided by a step-level value model that rewards moderate surprisal (curiosity) while penalizing incoherence, and Odds Ratio Preference Optimization (ORPO) to fine-tune the policy on high-value plot expansions.
This iterative reinforcement learning loop systematically explores multiple candidate plot branches, backpropagates quality signals, and adapts the policy for faster convergence, notably shifting the policy from puzzle-based Chain-of-Thought to more character-driven storytelling.
In small-scale tests with short-story prompts, 67\%--77\% of participants favored COS(M+O)S's highest-rated expansions over lower-rated ones, suggesting that our learned value function aligns.
GPT-4o ratings further show that COS(M+O)S surpasses naive single-pass decoding from Llama 3.2 3B by 0.59 SD, coming within 0.06 SD of Llama 3.1 70B (no significant difference, $p=\text{0.93}$).
Pairwise comparisons with o1 place COS(M+O)S 1.5 SD above the 3B baseline and find no statistically significant gap from 70B.
Nevertheless, absolute story quality remains modest, constrained by the small model's capacity and limited training data.
\end{abstract}

\section{Introduction}\label{introduction}

Recent advances in large language models (LLMs) have revealed their remarkable capacity for composing coherent prose, storytelling, and even tackling mathematical or coding problems. Yet LLM generators default to a single-pass System 1 style of reasoning \citep{kahneman_representativeness_2002}, generating story text (or solutions) in one shot based on autoregressive next-token predictions. This approach yields predictable or formulaic expansions that, in the context of plot quality, lack novelty and engagement. Indeed, under a standard cross-entropy training regime, the most frequent continuations from the training corpus tend to dominate, limiting creativity. Even with higher sampling temperatures, autoregressive decoding still draws from these archetypes.

\begin{figure}[H]
	\centering
	\includegraphics[width=\columnwidth]{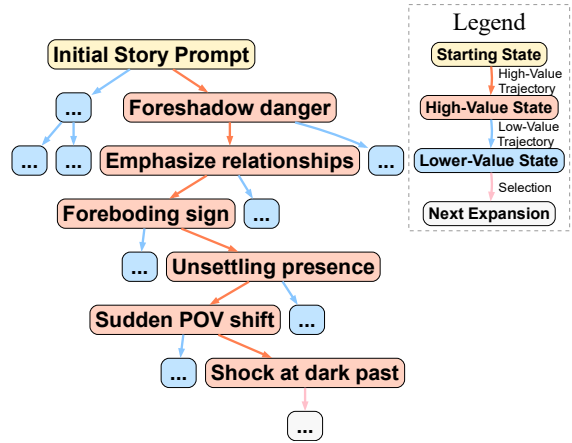}
	\caption{\textbf{MCTS-based story exploration and summarized (Chain-of-Thought) actions.} Each edge denotes a candidate action proposed by the policy model, and each node is a partial story state. High-value expansions (orange) are explored more deeply. After search concludes, actions with high $Q$ serve as evidence for ORPO fine-tuning.}
	\label{fig:fig1}
\end{figure}

A promising direction, inspired by test-time compute scaling \citep{snell_scaling_2024, qi_mutual_2024}, is to adopt a System 2 strategy that unfolds over multiple reasoning steps rather than in a single pass. In mathematical problem-solving, for example, recent works paired a policy LLM with a reward or value model in an iterative feedback loop \citep{openai_openai_2024, xie_monte_2024}, thereby refining solutions in a manner more akin to human deliberation. Analogously in story generation, we show that the same principle can be applied to expand the story step by step, selectively refining or discarding suboptimal plot branches with the help of a search algorithm and a fine-grained evaluation.

In this paper, we introduce a new System 2-inspired framework, \textbf{COS(M+O)S} (\underline{C}uriosity-\underline{O}riented \underline{S}tep-Level \underline{M}CTS+\underline{O}RPO \underline{S}trategy), to tackle open-ended storytelling with LLMs. While our experiments focus on the relatively compact Llama 3.2 3B for computational feasibility, it could equally be applied to larger backbones like Llama 3.1 70B \citep{grattafiori_llama_2024}. Rather than relying on a single pass, we introduce an iterative mechanism that leverages Monte Carlo Tree Search (MCTS) \citep{coulom_efficient_2007, kocsis_bandit_2006, browne_survey_2012} alongside a \emph{step-level} value model. In our implementation, Llama 3.2 3B serves both as a \emph{policy model} for proposing possible plot-focused Chain-of-Thought (CoT) "actions" at each step (reasoning how the story should progress) and as a \emph{simulation model} that applies the chosen CoT to generate the next story segment. The value model evaluates these segments. MCTS explores multiple candidate next actions, informed by a curiosity signal that rewards moderate surprisal (as a proxy for originality; intellectual engagement) and penalizes incoherence. Through repeated expansion and backpropagation within the search tree, we discover plotlines that strike a meaningful balance between formulaic familiarity and fresh intrigue (Figure \ref{fig:fig1}).

As in other System 2-style techniques, a key challenge is the overhead of iteratively generating and discarding large numbers of expansions. While MCTS can uncover high‐value trajectories, it often requires extensive search. To address this, we integrate Odds Ratio Preference Optimization (ORPO) \citep{hong_orpo_2024}, which we use to incorporate the partial expansions discovered by MCTS. Drawing on the concept of process-level (step-level) reward modeling \citep{lightman_lets_2023}, MCTS backpropagates the step-level value model signals, computing an action‐value $Q(s,a)$ for each expansion. Expansions with higher $Q$ are flagged as "preferred" examples in ORPO fine‐tuning, prompting the policy to internalize these higher‐value actions. This self-evolvable reinforcement learning loop---MCTS discovering good expansions guided by the value model, then fine-tuning the policy---demonstrably boosts story quality while reducing brute-force search demands.

Empirically, we show that even with a compact 3B-parameter backbone (Llama 3.2 3B), COS(M+O)S top MCTS expansions are favored over the lowest-value expansions by human participants 67\%--77\% of the time, closely matching GPT4o (validating our value model). Moreover, GPT-4o's \citep{openai_gpt-4o_2024} absolute quality ratings place these best expansions 0.59 SD above the naive baseline (no MCTS), and o1-based pairwise comparisons (plot preferences) reveal a 1.5 SD improvement over that same baseline. Thus, a smaller LLM (3B parameters) under our iterative search-and-fine-tune procedure closes much of the gap to naive single-pass decoding from bigger models, such as Llama 3.1 70B, validating that \emph{System 2}-style iterative reasoning and search can serve as a viable strategy in open-ended creative tasks. Interestingly, the synergy between curiosity-driven exploration, MCTS and policy refinement moves us beyond formulaic, puzzle-like expansions---introducing hidden motivations, interpersonal conflict, character-development, or subtle foreshadowing in the CoT reasoning (Figure \ref{fig:fig1}).

\paragraph{Contributions.} We summarize our primary contributions as follows:

\noindent (1) We introduce \textbf{COS(M+O)S}, a \emph{System 2}-inspired framework for story generation that leverages MCTS with an inverted-U curiosity index to systematically explore creative but coherent plot branches. 

\noindent (2) We couple MCTS with ORPO to internalize newly discovered "good" expansions, thus speeding up convergence to more engaging plots. 

\noindent (3) Through controlled experiments, we demonstrate that even with a smaller policy model (3B parameters), COS(M+O)S generates plots favored by both human and automated evaluations, offering a scalable path to higher-quality text generation. 

\section{Related Work}\label{sec:related_work}

MCTS has demonstrated its effectiveness across multiple decision-making domains, from the seminal work in AlphaGo \citep{silver_mastering_2016} and AlphaZero \citep{silver_mastering_2017} to mathematical problem-solving \citep{trinh_solving_2024}, and robotics tasks under partial observability \citep{garrett_online_2020}. In text generation, Yao et al. \citep{yao_plan-and-write_2019} proposed a plan-and-write framework that creates a plot outline before expanding it into a full narrative, but it operates primarily in a single pass rather than through iterative search. In parallel, reward-shaping approaches have been used to control neural story plot generation toward specific goals \citep{tambwekar_controllable_2019}. Meanwhile, the notion of \emph{surprisal} from psycholinguistics \citep{hale_probabilistic_2001, levy_expectation-based_2008} has motivated curiosity-driven exploration in reinforcement learning \citep{burda_large-scale_2018, le_beyond_2024}, though for game environments rather than stories. More recent MCTS-based methods extend into multi-step language, math, or code generation, \citep{chen_alphamath_2024, xie_monte_2024, feng_alphazero-like_2024} sometimes employing reward modeling or preference optimization \citep{zelikman_star_2022, qi_mutual_2024, zhao_marco-o1_2024, bai_training_2022, rafailov_direct_2024}. Complementary "slow thinking" \citep{min_imitate_2024} and "self-refine" \citep{madaan_self-refine_2023} strategies iterate on an LLM's CoT for logical tasks, while OpenAI's o1 model \citep{openai_openai_2024} integrates CoT reasoning with advanced search, substantially boosting performance on competition math and coding tasks. Recent work also shows how a policy trained with PPO can guide MCTS decoding for tasks like toxicity mitigation \citep{liu_dont_2024}.

\section{Methods}\label{sec:method}

\paragraph{Overview of the Proposed Framework}

\noindent COS(M+O)S integrates a policy model, a simulation model, and a step-level value model within the MCTS framework. The policy model $\pi(a|s)$ proposes candidate plot actions at a given story state $s$, the simulation model uses these actions to advance the story (Section \ref{sec:policy_model}). The step-level value model $V(s)$ (Section \ref{sec:value_model}) evaluates the quality of the resulting plots. By iteratively applying MCTS, we explore a large space of potential stories, balancing exploration of new plot branches with exploitation of promising ones. We then apply ORPO to fine-tune the policy model, using MCTS-derived preferences as training signals.

\subsection{Monte Carlo Tree Search (MCTS)}\label{sec:mcts}

We treat plot development as a sequential decision process represented by a tree $\mathcal{T}$. Each node $s$ corresponds to a partial (or fully generated) story state, and edges represent possible plot-expanding actions $a \in \mathcal{A}(s)$. Fully generated stories appear at final-depth states $\mathcal{L}_f$, from which we can define
\begin{equation}
	\label{eq:v_max}
	V_{\max}^{(\mathrm{final})}
	\;=\; 
	\max_{s \,\in\, \mathcal{L}_f} \, V(s),
\end{equation}
the highest estimated plot quality found during search.

\noindent We employ a standard MCTS procedure \citep{kocsis_bandit_2006,browne_survey_2012} with five phases per iteration:

\noindent\textbf{(1) Selection} [using the Upper Confidence Bound, UCB],\\
\textbf{(2) Expansion} [sampling new actions from the policy $\pi(a|s)$],\\
\textbf{(3) Simulation} [applying the chosen action to advance the story],\\
\textbf{(4) Evaluation} [using our step-level value model $V(s)$, deferred until at least 50\% of the story is generated], and \\
\textbf{(5) Backpropagation} [updating action-value estimates $Q(s,a)$ along the path. For each state-action pair $(s,a)$ along the path from child $s'$ up the tree, we apply Equations \ref{eq:NsUpdate}--Equation \ref{eq:QsaUpdate}: 
\begin{align}
	\label{eq:NsUpdate}
	N(s,a) &\leftarrow N(s,a) + 1, \\
	\label{eq:WsUpdate}
	W(s,a) &\leftarrow W(s,a) + V(s'), \\
	\label{eq:QsaUpdate}
	Q(s,a) &\leftarrow \frac{W(s,a)}{N(s,a)}.
\end{align} 
Here, $W(s,a)$ is the cumulative value for action $a$ at state $s$, and $N(s,a)$ is the visit count.]

\noindent Notably, our step-level approach assigns "partial credit" to unfinished plots, enabling MCTS to explore promising half-complete branches. We incorporate several additional strategies (e.g., a high initial branching factor and parallel expansions) to efficiently navigate large story spaces; see Appendix \ref{app:algorithm} for complete details and hyperparameters.

\subsection{Policy and Simulation Models}\label{sec:policy_and_simulation_model}

\paragraph{Policy Model and ORPO Optimization}\label{sec:policy_model}

\noindent We employ a language model (Llama 3.2 3B) to serve as our policy $\pi(a|s)$ that generates possible plot actions $a$ given the current story state $s$ (akin to a AlphaZero approach). To elicit plot actions (CoTs) in a structured format, we prompt our \emph{base policy model} $\pi_0(a|s)$ with a fixed template (sampling temperature of 0.7). Initially, MCTS exclusively uses $\pi_0(a|s)$ to propose actions at each step.

After each round of MCTS searching (Section \ref{sec:experiments}), we apply ORPO to fine-tune the policy. Specifically, we leverage the action-value estimates $Q(s,a)$ (obtained via the value model and backpropagation; Equation \ref{eq:QsaUpdate}) to label actions as "chosen" (i.e., higher-quality) or "rejected" (lower-quality). During ORPO, we increase the probability that the policy selects chosen actions while decreasing the probability of rejected ones, progressively approximating an optimal policy $\pi^*$. This process aligns the policy $\pi_{\text{trained}}(a|s)$ toward higher-value trajectories uncovered by MCTS.

To retain the diversity of the base policy model while progressively incorporating improvements, we mix actions proposed by $\pi_0(a|s)$ and $\pi_{\text{trained}}(a|s)$ at a 50:50 ratio in subsequent rounds. Further implementation details, including dataset generation and training hyperparameters, appear in Appendix \ref{app:orpo_details}.

\noindent \paragraph{Simulation Model}\label{sec:simulation_model}

\noindent The simulation model uses the Llama 3.2 3B language model (with temperature set to 0.0) to deterministically generate story continuations. For each newly expanded node, the simulation model is given the entire story trajectory, along with the selected next action and previous CoTs. Using these inputs, it produces the next story state (plot event).

We use $K=\text{32}$ bullet points to condense the story's major developments into manageable steps, ensuring that our surprisal-based metrics highlight pivotal plot transitions rather than minor textual details. The bullet-point format also ensures consistent chunking and guides the LLM to produce coherent story segments at each step.

\subsection{Step-Level Value Model}\label{sec:value_model}

We define $V(s)$ as our quality measure for a story state $s$. A calibrated Support Vector Classifier (SVC) distinguishes plots as "good" or "bad" based on PCA-transformed features, outputting a probability that the story is "good". We adopt this probability as $V(s) \in [0,1]$. Higher values indicate a greater likelihood of high plot quality under our chosen metrics, while we recognize that true story quality remains inherently subjective.

\paragraph{Curiosity Index}

\noindent To approximate intellectual engagement (novelty), we introduce a \textit{curiosity index} based on token-level surprisal computed using the Phi-3.5 3B model \citep{abdin_phi-3_2024}.

\begin{figure}[htb]
	\centering
	\includegraphics[width=0.85\columnwidth]{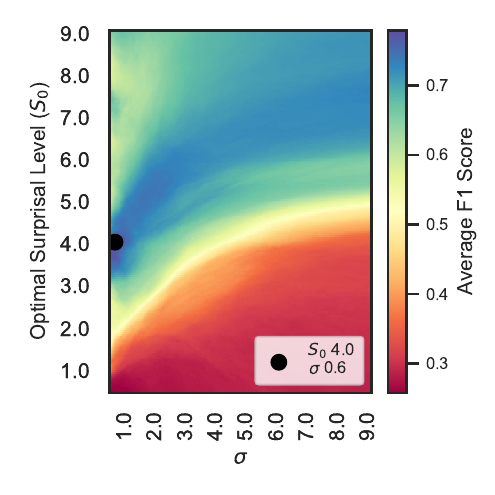}
	\caption{Heatmap of average F1 score for separating stories via the curiosity index, using group-aware stratified repeated k-fold cross-validation. The optimal surprisal peak shifts with model size: a smaller, "less read" SmolLM-360M \protect\citep{allal_smollm_2024}) exhibit a higher "optimal" surprisal level (around 10 bits), reflecting that it finds these stories more surprising than the bigger Phi-3.5 3B model.}
	\label{fig:hyperparameter_curiosity}
\end{figure}

Surprisal for each token $i$ is 
$
S(i) = -\log_2 P(i \mid \text{context}_{<i})
$
, where $P(i \mid \text{context}_{<i})$ is the token probability given its preceding context. We map $S(i)$ to an \textit{interest function} $I(i)$ inspired by the (inverted-U) Wundt curve \citep{bianchi_magic_2014}:
\begin{equation} 
	\label{eq:interest_function}
	I(i) = e^{-\frac{(S(i) - S_0)^2}{2 \sigma^2}}.
\end{equation}
In Equation \ref{eq:interest_function}, $S_0$ represents the optimal surprisal level, and $\sigma$ controls the spread of the engagement peak. We set $S_0=\text{4}$ bits and $\sigma=\text{0.6}$ bits based on hyperparameter tuning (Figure \ref{fig:hyperparameter_curiosity}). Averaging $I(i)$ over all tokens yields the curiosity index.
Figure \ref{fig:distribution_curiosity} in the Appendix \ref{app:value_model_details} shows the distribution of the curiosity index for both classes.

\paragraph{Value Model Training}

\noindent We train the value model to distinguish stories using features derived from the curiosity index, semantic coherence, and other measures. We, for example, broaden coverage of the "bad" class by including stories from smaller LLMs, whose reduced parameter counts should degrade story quality (GPT-3.5 \citep{brown_language_2020}, Llama 3 7B \protect\citep{grattafiori_llama_2024}, Gwen 2 7B \citep{yang_qwen2_2024}, and Mixtral 8x7B \citep{jiang_mixtral_2024}).

All plots are normalized into a bullet point list of $K=\text{32}$ elements using GPT-4o, and we evaluate them at multiple completion levels (50\%, 63\%, 75\%, 88\%, and 100\% of the intended length) to capture evolving story quality. As mentioned above, our simulation model likewise produces bullet-point plots in a matching format; stylistic differences arise between GPT-4o-generated transformations and our simulation model. 

We preprocess these extracted features via imputation, standardization, and PCA, then optimize hyperparameters via group-aware stratified repeated k-fold cross-validation. Notably, $V(s)$ closely tracks $Q(s, a)$, making random rollouts unnecessary. For details, please refer to Appendix \ref{app:value_model_details}.

\section{Experiments and Results}

\subsection{Experimental Setup}\label{sec:experiments}

We evaluated our proposed framework on short-story generation tasks, using 18 randomly selected story prompts, given our resource constraints. All MCTS experiments were conducted on a single consumer GPU.

We ran MCTS for 100 iterations per prompt. The policy and simulation models each used a quantized (Q5\_K\_M) Llama 3.2 3B backbone, with hyperparameters and training details described in Section \ref{sec:policy_and_simulation_model}. We set the MCTS parameters (e.g., exploration constant and dynamic branching) according to Section \ref{sec:mcts} and Appendix \ref{app:algorithm}, and used identical deferred evaluation thresholds (50\% of story length) across all runs.

After each round of runs, we applied ORPO to refine the policy model. This process was repeated for two rounds, each with new prompts (i.e., a held-out test set). Where needed, implementation and hyperparameter details---e.g., for the SVC, curiosity index, and coherence score---appear in Appendices \ref{app:orpo_details} and \ref{app:value_model_details}.

\subsection{Results}\label{sec:results}

\paragraph{Value Model Performance}

\noindent We first evaluate the value model's ability to distinguish "good" and "bad" stories. Table \ref{tab:classification_report} shows detailed classification metrics on the held-out test set. The model achieves a macro-average F1 score of 0.82. 

\begin{table}
	\centering
	\caption{Classification metrics (precision, recall, F1) on the test set for "good" vs. "bad" stories, evaluated on stories with $\ge$ 50\% completion.}
	\begin{tabular}{lccc}
		\toprule
		\textbf{Class} & \textbf{Precision} & \textbf{Recall} & \textbf{F1-Score} \\
		\midrule
		bad  & 0.81 & 0.85 & 0.83  \\
		good & 0.84 & 0.80 & 0.82  \\
		\midrule
		\textbf{Macro Average} & 0.83 & 0.82 & 0.82 \\
		\bottomrule
	\end{tabular}
	\label{tab:classification_report}
\end{table}

Appendix \ref{app:results-svc}, we visualize the SVC's decision boundaries using the first two and three principal components, present the confusion matrix, and show the value model's accuracy as a function of how much of the story has been generated (Figure \ref{fig:model_accuracy_over_text_portion}). These results confirm that accuracy improves with plot progression.

\paragraph{Effectiveness of MCTS and Policy Model Refinement}\label{sec:mcts-orpo-results}

\noindent We hypothesized that our MCTS framework would improve plot quality through iterative exploration alone, and that further preference alignment via ORPO would accelerate these gains. To test this, we conducted three rounds of experiments, each comprising six separate MCTS runs initialized with different story prompts, resulting in just 18 stories in total.

In \textbf{Round 0}, MCTS used the base (untrained) policy to propose actions during MCTS. After collecting each action's $Q$-value, we applied ORPO fine-tuning to form the \textbf{Round 1} policy. We then repeated MCTS on fresh prompts, gathered new data, performed another round of ORPO, and obtained the \textbf{Round 2} policy. Throughout, we measured final-depth story quality via $V_{\max}^{(\mathrm{final})}$ (Section \ref{sec:mcts}; Equation \ref{eq:v_max}). Notably, each new round used a fresh set of initial prompts (randomly chosen), ensuring that the fine-tuned policy was evaluated on out-of-distribution story contexts rather than reusing the same data from previous rounds. For example, one of the initial story prompts is shown in Text Box \ref{textbox:initial-story}.

\begin{textbox}\label{textbox:initial-story}
	- Begin your story with the sensation of a breeze brushing against a character's skin.
	
	- Set your story on a sailboat, large or small, and entirely at the mercy of the winds.
	
	- Craft a story where a character finds themselves swept away by the wind, either physically or metaphorically.
\end{textbox}

\begin{figure}[!htb]
	\centering
	\includegraphics[width=0.95\columnwidth]{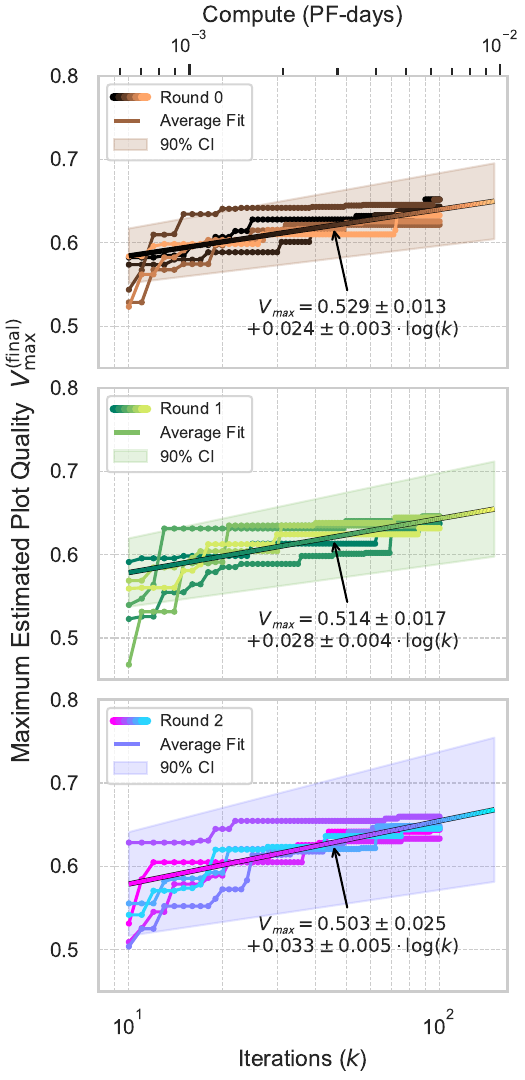}
	\caption{Progression of maximum estimated plot quality $V_{\max}^{(\mathrm{final})}$ as a function of iterations $k$ and compute (PF-days) in Round 0 (base policy), Round 1 and Round 2 (fine-tuned policies). Individual experiments are plotted alongside the average fit and its 90\% confidence interval. We estimate PF-days by multiplying the run time by 35 teraflops for a GPU operating at an assumed 30\% utilization.}
	\label{fig:policy_round0}
\end{figure}

\begin{figure}[htb]
	\centering
	\includegraphics[width=0.95\columnwidth]{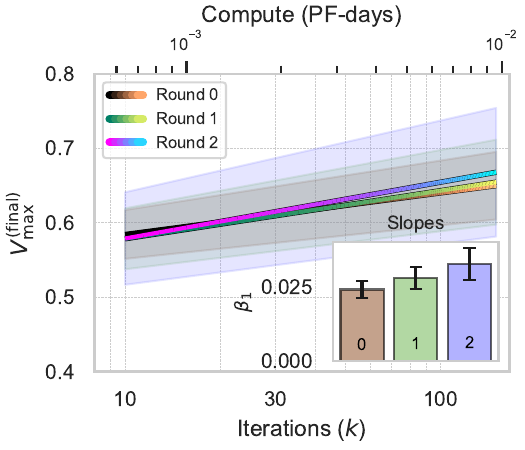}
	\caption{Log-linear fits (including 90\% CI) of the average maximum estimated plot quality $V_{\max}^{(\mathrm{final})}$ across Rounds 0, 1, and 2 on a log scale of MCTS iterations $k$ (and corresponding PF-days of compute).}
	\label{fig:policy_summary}
\end{figure}

Figure \ref{fig:policy_round0} shows the evolution of the maximum estimated plot quality $V_{\max}^{(\mathrm{final})}$ against MCTS iterations $k$, revealing three key results. First, \emph{Round 0} (no ORPO) already improved story quality through systematic search over the space of CoTs alone; the growth rate is roughly log-linear in $k$. Second, \emph{Rounds 1--2} discovered high-value expansions faster; see how their curves rise more steeply (i.e., with fewer iterations) for the same $V_{\max}^{(\mathrm{final})}$. Third, we confirmed this effect via a mixed-effects linear regression to assess how $V_{\max}^{(\mathrm{final})}$ scales with $\log(k)$ while accounting for variability across different experiments:
\begin{equation} 
	V_{\max}^{(\mathrm{final})} = \beta_0 + \beta_1 \cdot \log(k) + u_g
\end{equation}
where $u_g$ represents random intercepts and slopes. We find $\beta_1 >$ 0 $(p <\text{0.001})$ within the tested iteration range ($k=$ 1 to $k=$ 100) for all rounds but with increased $\beta_1$ from Round 0 to Round 2, indicating that fine-tuned policies discover higher-value paths more rapidly. Figure \ref{fig:policy_summary} consolidates these fits by plotting the average log-linear trajectories for Rounds 0--2, demonstrating an upward shift in each successive round. Consistently, the final mean $V_{\max}^{(\mathrm{final})}$ increased from 0.638 $\pm$ 0.012 in Round 0 to 0.641 $\pm$ 0.005 in Round 1 and 0.646 $\pm$ 0.008 in Round 2, corresponding to gains of 0.25 and 1.0 baseline SD, respectively (Appendix \ref{app:rel-improvement}).

To quantify speedups, we measured how many iterations each round needed to exceed either a 10\% or a 20\% gain in $V_{\max}^{(\mathrm{final})}$, relative to the earliest-completion reference (i.e., $k = 8$, the first iteration at which a story is fully generated). For a 10\% improvement, ORPO-fine-tuned policies in Round 1 and Round 2 reached that threshold 1.4$\times$ and 1.5$\times$ sooner than Round 0. For a 20\% improvement, Round 1 and Round 2 showed 1.6$\times$ and 1.9$\times$ speedups, respectively (Appendix \ref{app:rel-improvement}). 

\paragraph{Visualization of the Monte Carlo Tree Search Process}

\begin{figure}[htb]
	\centering
	\includegraphics[width=\columnwidth]{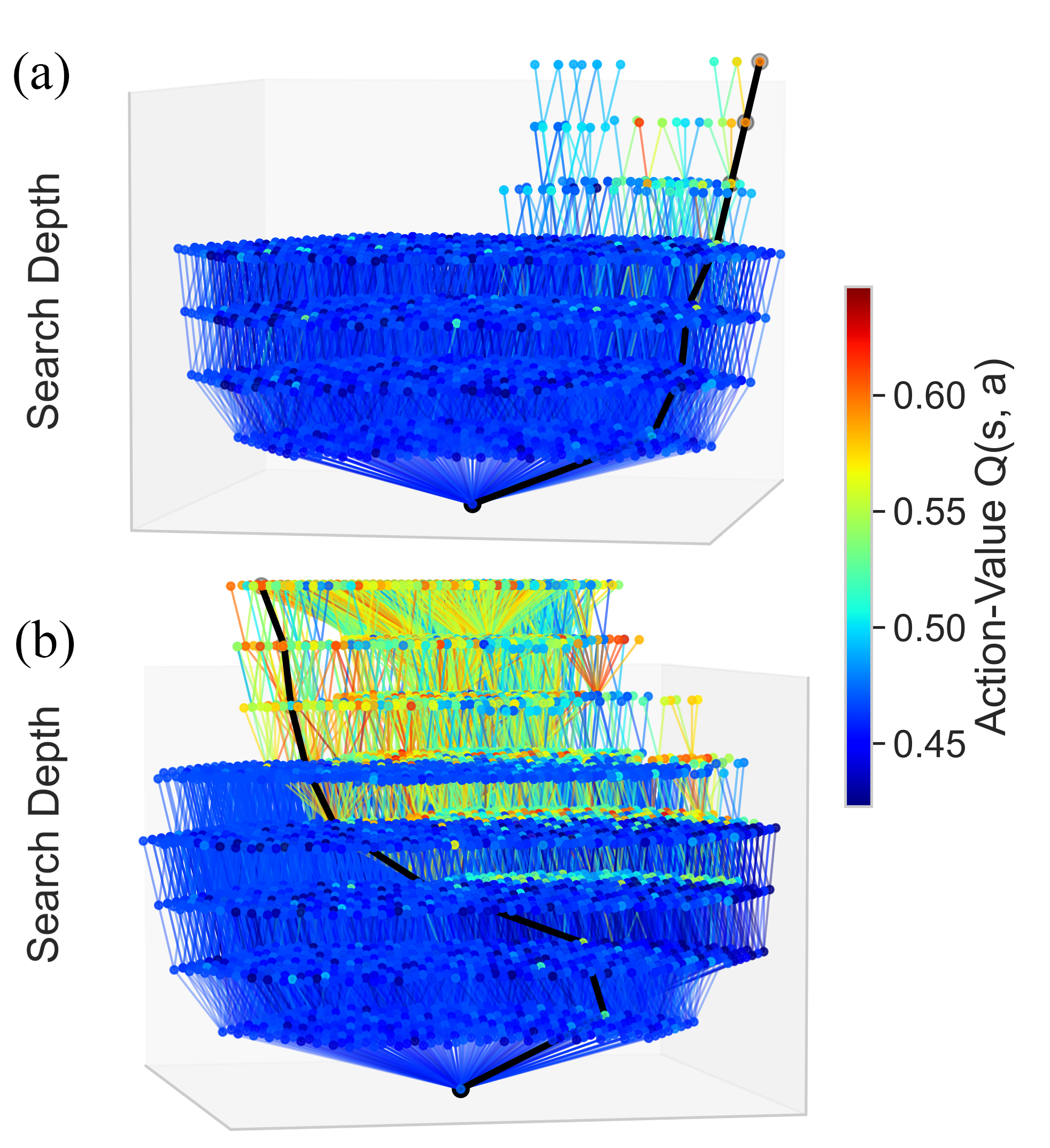}
	\caption{Visualization of the MCTS search tree of an experiment in Round 0 after 8 (a, top) and 80 (b, bottom) iterations. Dots represent story states, with the vertical axis denoting search depth and the horizontal axes serving only for layout. Colors correspond to action-value estimates $Q(s,a)$ (blue = lower, red = higher). The black line highlights the highest-value trajectory $V_{\max}^{(\mathrm{final})}$.}
	\label{fig:tree-evolution}
\end{figure}

\noindent Figure \ref{fig:tree-evolution} shows the MCTS tree at 8 and 80 iterations, with nodes representing story states and the estimated action values $Q(s,a)$. Early on (Figure \ref{fig:tree-evolution}a), the search is shallow and broad, with primarily lower $Q$ values. By iteration 80 (Figure \ref{fig:tree-evolution}b), MCTS has focused on the most promising trajectories, culminating in a clearly defined highest-value path (black line) $V_{\max}^{(\mathrm{final})}$. Branches that remain unexplored or unpromising retain lower $Q$ values.

\paragraph{Human and Model-Based Evaluations}\label{sec:human-model-evals}

\noindent Below, we compare the best (highest-value) and worst (lowest-value) complete stories from MCTS to validate whether our estimated plot quality $V_{\max}^{(\mathrm{final})}$ aligns with external criteria. We also evaluated how our COS(M+O)S approach (i.e., highest-value MCTS) compares to a naive baseline (no MCTS; single-pass) of several Llama model variants (at temperature 0.7). Specifically, we aggregated 50 repeated GPT-4o ratings per story (18 stories) into a single mean rating per story.

Notably, the MCTS process had already filtered out obviously poor candidates using our selection strategies (e.g., UCB, Appendix \ref{app:algorithm} Equation \ref{eq:ucb}). Consequently, the lowest-value successful trajectories (completed stories) are not inherently bad, on average they have $V_{\min}^{(\mathrm{final})}=0.52$ vs. $V_{\max}^{(\mathrm{final})}=0.64$ for the top trajectories. From Section \ref{sec:mcts}, the successful highest-value trajectory is that which maximizes $V(s)$ among fully generated plots, while the lowest-value trajectory is that which minimizes $V(s)$, i.e., $V_{\min}^{(\mathrm{final})} = \min_{s \in \mathcal{L}_f} V(s)$. These labels strictly reflect MCTS's internal estimates, see Figure \ref{fig:tree-evolution}.

\paragraph{A/B Preference Test with Human Participants} 

\noindent We recruited 8 participants to rate 4 story pairs; on average, 8 participants responded per pair. In total, we collected 30 pairwise preferences, and 67\%--77\% of these favored the highest-value plot, suggesting our value model aligns with human judgment. Detailed methodologies for these evaluations are provided in Appendix \ref{app:ab-comparison}.
\begin{figure*}[htb]
	\centering
	\includegraphics[width=\textwidth]{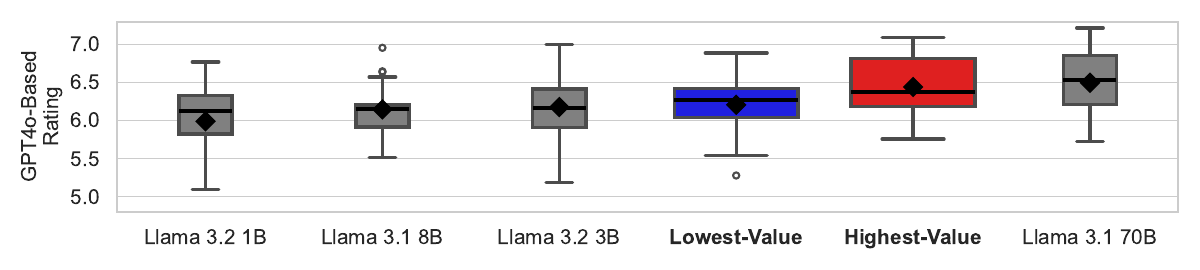}
	\caption{Distribution of GPT-4o's aggregated (1-10) ratings across nine literary dimensions, comparing highest- and lowest-value MCTS trajectories with Llama System 1-baselines (no MCTS). Llama 3.1 70B surpasses Llama 3.2 3B by 0.64 SD, while COS(M+O)S leads Llama 3.2 3B by 0.59 SD. The x-axis is ordered by the mean story rating from 18 stories (50 GPT-4o evaluations for each story).}
	\label{fig:gpt4o-eval}
\end{figure*}

\paragraph{GPT-4o Evaluation} 

\noindent We asked GPT-4o to rate each story's plot quality (1--10) across nine dimensions (plot structure, tension, originality, character development, overall impact, theme, conflict, pacing, style/voice, details in Appendix \ref{app:ab-comparison}). We used an \emph{absolute (non-comparative)} rating setup, where GPT-4o evaluated each story in isolation. This design reduces direct comparison bias but can compress scores into a narrower band since GPT-4o is not assessing stories side by side. Figure \ref{fig:gpt4o-eval} shows that our highest-value MCTS trajectories exceed the lowest-value ones by about 0.65 SD (paired Wilcoxon $p = \text{0.01}$, paired Cohen's $d = \text{0.57}$). Under a common-language effect-size measure \citep{mcgraw_common_1992}, $d = \text{0.57}$ implies $\approx$ 66\% probability that a random draw from the higher-rated set surpasses one from the lower-rated set, closely matching the 67\%--77\% preference observed in our human A/B study (Section \ref{sec:human-model-evals}).

Beyond lowest-value trajectories itself, COS(M+O)S outperformed Llama 3.2 3B (no MCTS) by 0.59 SD on GPT-4o's scale (borderline at $p=\text{0.09}$, $d = \text{0.45}$). Strikingly, Llama 3.1 70B is only 0.06 SD above COS(M+O)S (Wilcoxon $p = \text{0.93}$, $d=\text{0.06}$), indicating no difference at conventional significance levels.

These results highlight that our method greatly enhances Llama 3.2 3B's GPT4o-evaluated plot quality---closing much of the gap to Llama 3.1 70B. Crucially, GPT-4o's independent ratings validate our internal value model's assessments, reinforcing the model's ability to capture key facets of plot quality recognized by external evaluations.

\paragraph{o1-Based Ratings} 

\noindent Additionally, we performed pairwise plot-quality evaluations with o1 \citep{openai_openai_2024} on a 0--100 scale (Appendix \ref{app:o1-comparison-results}). As shown in Table \ref{tab:o1-comparison-results-llama3b}, COS(M+O)S surpassed the naive 3B baseline by 1.5 SD (paired $t$-test, $p < \text{0.05}$). In a separate comparison (Table \ref{tab:o1-comparison-results-llama70b}), it outscored Llama 3.1 70B by 0.4 SD, though this was not statistically significant ($p = \text{0.36}$).

\section{Discussion}\label{sec:discussion}

Our results show that combining MCTS with preference-based fine-tuning (ORPO) can improve plot quality for a smaller 3B-parameter model, narrowing the gap to a 23$\times$ larger 70B model on our short-story tasks. Rather than relying on a single-pass strategy, the MCTS approach systematically explores multiple plot branches, and ORPO progressively internalizes the best expansions for faster convergence. Qualitative and quantitative evaluations (including a small A/B test with 8 participants on 4 prompts, in which 67\%--77\% favored MCTS's top trajectories, plus GPT-4o- and o1-based ratings) confirm these improvements, suggesting that System 2---inspired iterative techniques can act as another improvement dimension. At the same time, we observe a log-linear relationship between the number of MCTS iterations and quality gains in our tested regime, meaning significant computational overhead as stories grow in length. On these limited test prompts, the 3B approach performed comparably to the 70B baseline. The demonstrated synergy between MCTS-driven exploration and fine-tuned policy refinement offers a promising path for generating more engaging, novel plots even under constrained model sizes.

Notably, we achieve these improvements using a small quantized language model (SLM) for both policy and simulation. Qualitatively, we observe a marked transition away from abstract-, \text{puzzle-,} or object-based archetypes in the base policy toward more interpersonal twists, emotional confrontations, hidden motivations, and other forms of character-driven storytelling. This shift suggests that our cognitively inspired features---e.g., the curiosity index and coherence score---meaningfully encourage more engaging storylines. We emphasize, though, that these metrics may undervalue genres favoring lower surprisal (e.g., comfort fiction, cozy stories).

Despite promising outcomes, our framework has practical constraints. We tested on a relatively small language model (3B parameters), while a larger backbone might yield further emergent gains. We also depend on the model's context window for logical consistency, lacking explicit reference tracking. Partial evaluations can drift if the reward model is imperfectly aligned, risking reward hacking. Although MCTS excels in well-defined tasks such as code or math \citep{zhao_marco-o1_2024, trinh_solving_2024}, story generation lacks a single correct outcome---complicating reward signals and risking distribution shifts. User preference methods \citep{leike_scalable_2018} can help but often demand extensive human or computational resources, and deeper expansions \citep{snell_scaling_2024, brown_large_2024} may exacerbate misalignment in subjective contexts. More robust or multi-pronged evaluators, or domain-specific constraints (e.g., historical accuracy for period pieces), could mitigate these risks. Our results demonstrate that combining a cognitively motivated value model with MCTS and preference-aligned fine-tuning can yield notably more compelling stories, even under limited compute resources.

\section{Conclusion}\label{sec:conclusion}

We presented \emph{COS(M+O)S} as a search-based framework integrating MCTS-guided CoT generation, step-level value modeling, and ORPO-based reinforcement learning to enhance LLM-driven storytelling. By using step-level evaluations after each expansion, we uncovered higher-value plot expansions. We showed that MCTS alone improved plot quality but scales only log-linearly with compute; by fine-tuning the policy on MCTS-discovered preferences, we accelerated convergence and reduce brute-force search costs. Nevertheless, these are only relative improvements, and the absolute story quality remains modest, limited by the small model size and training data. Evaluations with human readers, GPT-4o, and o1 confirmed that COS(M+O)S outperforms naive decoding on our test set. Future work can investigate larger backbones, domain-specific rewards, or explicit reference-tracking to further enhance long-form story coherence.

\subsection{Limitations and Future Directions}

\paragraph{Model size:} We rely on 3B-parameter models for both policy and simulation due to limited compute resources. Using far larger models (e.g., 70B+ parameters) would boost story quality and may reveal emergent capabilities, but could also complicate the log-linear improvement pattern. We might need careful hyperparameter tuning to maintain stable gains at larger scales.

\paragraph{Value model coverage:} Our surprisal- and coherence-based metrics are aimed to approximate cognitive processes but do not capture deeper literary elements such as thematic development or nuanced characterization. We have not fully dissected which features drive our value model's predictions, and it may undervalue genres favoring low-surprise plots.

\paragraph{Generative biases:} The base policy tends to produce puzzle-like tropes, likely reflecting its pretraining corpus. Without deeper data transparency, we cannot precisely diagnose these biases.

\paragraph{Human-centric evaluation:} Our human preference tests and GPT-4o ratings suggest meaningful quality improvements, yet we only sampled a small group and used 18 short-story prompts. These were chosen to ensure manageable compute while covering a few diverse topics, but a larger-scale study with more prompts and participants would offer stronger evidence of generalization. Future studies could expand the participant pool and evaluate nuanced dimensions of reader immersion.

\paragraph{Caveats on "Closing the Gap" to 70B:} Although our GPT-4o and o1-based evaluations sometimes place COS(M+O)S close to or even slightly above naive Llama 3.1 70B in overall scores, these comparisons rely on a relatively small sample of prompts and often do not reach conventional statistical significance. In addition, both GPT-4o and o1 are themselves LLM-based evaluators, which can introduce unknown biases. Our human preference study, while encouraging, involved only eight participants across a few story pairs, so the results should be viewed as suggestive rather than definitive. Larger-scale, more diverse human evaluations would be needed to robustly confirm the extent to which smaller LLMs truly match or surpass naive 70B outputs on creative-writing tasks.

\paragraph{Reward hacking and reference tracking:} Our policy model could learn shortcuts that boost $Q$-values without delivering truly coherent plots. Because we rely on the model's attention, and context window, suprisal-based features and coherence scoring, partial expansions can inadvertently produce contradictory states (e.g., inconsistent character references). We do not currently track references explicitly. A multi-pronged evaluation setup, a reference-tracking system, or a knowledge graph would help prevent such loopholes and maintain consistency.

\paragraph{Search complexity and scalability:} We explored only short stories. MCTS becomes expensive as story length (search depths) grows. Adopting hierarchical expansions, better parallelization, or more efficient tree-search heuristics could make the method more scalable for longer stories. A larger LLM backbone (e.g., 70B parameters) may also reduce the branching required since higher-quality expansions would emerge earlier.

\paragraph{Generalized value modeling:} While we rely in this study on a binary "good vs. bad" plot classification grounded in surprisal and coherence features, the MCTS+ORPO loop naturally accommodates different or more specialized evaluators. In principle, any scalar-valued "score" can serve as the value function, allowing the framework to tackle tasks far beyond narrative quality. For instance, one could incorporate domain-specific metrics---such as functional correctness for code generation (by executing candidate solutions), factual accuracy via external knowledge lookups (e.g., financial data or historical records), or step-by-step verification of mathematical proofs---to reward and propagate actions that meet these objective criteria. Similarly, subjective or multi-faceted goals (like style matching, educational appropriateness, or authenticity in song-writing) could be encoded through a carefully designed value model that balances competing priorities. In all such cases, the MCTS process remains unchanged: it proposes branching expansions, evaluates them using the new domain-specific value model, and backpropagates quality signals. By simply swapping in a different evaluator or combining multiple evaluators (e.g., one for correctness and one for creativity), the same iterative search-and-refine mechanism can be repurposed for a wide variety of multi-step reasoning and generation tasks, without altering the underlying MCTS+ORPO infrastructure.

\paragraph{Participant consent and potential misuse:} In our small-scale preference study (eight adult volunteers), participants gave informed consent and were not asked for any personal or sensitive data. Although our focus is on creative storytelling, the method itself does not include content moderation or bias mitigation. As a result, there is a risk that, in other contexts, it could be misused to generate disinformative or offensive material or reproduce stereotypes from its underlying language model. Future work might incorporate robust content filtering or bias control to address such concerns.

\section*{Acknowledgments}

I am sincerely grateful to all who generously took part in the A/B preference tests: Artemi Bendandi, Clarisse Faure, Danna De Boer, Florian M\"uller-Plathe, Han Dittmar, Kamonthira Wichai, and Lukas Materzok (listed in alphabetical order). A special thanks goes to Danna for proofreading the manuscript and suggesting improvements to its grammar and clarity.

\appendix 

\section*{\huge Appendices}

These Appendices detail our implementation strategies, data processing pipelines, and evaluation procedures. Section \ref{app:algorithm} describes the specific algorithmic strategies for Monte Carlo Tree Search (MCTS) and clarifies how we manage partial and fully generated stories. Section \ref{app:orpo_details} describes our fine-tuning setup and how we produce action-level preferences for Odds Ratio Preference Optimization (ORPO). Section \ref{app:value_model_details} provides additional details for the value model, and Section \ref{app:results-svc}, we visualize our Support Vector Classifier (SVC) decision boundaries for "good" vs. "bad" plots. Moreover, Section \ref{app:rel-improvement} highlights details of the MCTS+ORPO synergy. Finally, Section \ref{app:ab-comparison} provides methodological details for our external plot evaluations, including the A/B preference tests.

\section{Detailed MCTS Implementation and Algorithmic Strategies}\label{app:algorithm}

\paragraph{Tree Structure and Representation}

\noindent We represent the plot development process as a tree $\mathcal{T}$. Each node $s \in \mathcal{S}$ corresponds to a partial (or fully generated) story state, containing all plot content generated so far. In other words, each node embodies the entire text from the beginning up to that node's depth in the story; deeper nodes represent progressively longer story prefixes. A leaf node might be a fully written story (maximum depth) or a partially written one that is not (yet) further expanded. 

Each edge $a \in \mathcal{A}(s)$ represents a possible plot development, transitioning from one story state $s$ to another $s'$. 

\paragraph{MCTS Algorithm Steps}

\noindent Our Monte Carlo Tree Search procedure \citep{kocsis_bandit_2006,browne_survey_2012} consists of five main steps each iteration:

\noindent \textbf{Selection:} Starting from the root, we select child nodes that maximize the Upper Confidence Bound (UCB) criterion (Equation \ref{eq:ucb}):
\begin{equation}\label{eq:ucb}
	a^{*} = \arg\max_{a} \left[ Q(s,a) + c\sqrt{\frac{\ln N(s)}{N(s,a)}} \right],
\end{equation}
where $Q(s,a)$ is the current action-value estimate, $N(s)$ is the visit count of node $s$, $N(s,a)$ is the visit count of action $a$ at state $s$, and $c$ is the exploration parameter.

\paragraph{Expansion:} When we select a node $s$ that allows further exploration, we generate new actions $a$ using the policy model $\pi(a|s)$. Each selected action leads to a new child node $s'$, growing $\mathcal{T}$.

\paragraph{Simulation:} For each newly expanded node $s'$, the simulation model produces the next segment of the story by applying the chosen action $a$ to the current story state. This step advances the plot.

\paragraph{Evaluation:} We evaluate the resulting plot at node $s'$ using the step-level value model $V(s')$. To avoid unreliable early estimates, we defer all evaluations until the node's depth reaches 50\% of the intended total story length \citep{lightman_lets_2023,yin_reasoning_2024}. Once past that threshold, each new expanded node is scored by $V(s')$.

\paragraph{Backpropagation:} We propagate the evaluation $V(s')$ back up the tree to update action-value estimates. For each node-action pair $(s,a)$ along the path from $s'$ up the tree, we apply Equations \ref{eq:NsUpdate}--Equation \ref{eq:QsaUpdate}.

\paragraph{Algorithmic Strategies and Details}

\noindent While the standard MCTS framework above captures the core process, we incorporate several additional strategies and hyperparameters to handle the large branching factor and partial evaluations in story generation:

\paragraph{Initialization:} We initialize the tree $\mathcal{T}$ with root nodes containing initial story prompts. Parameters such as the maximum number of iterations $N$, the maximum depth of the tree $D$, initial maximum actions per state $\kappa$, and exploration parameter $c$ are defined.

\paragraph{Early Broad Exploration:} In the initial iteration, we focus on exploring a diverse range of plot paths by generating a large number of candidate actions at each node (e.g., $\kappa=\text{300}$). This broad exploration is meant to reduce the risk of missing promising story trajectories. Subsequent iterations reduce $\kappa$ to focus computational resources on the most promising paths.

\paragraph{Deferred Evaluation:} As described above, we do not evaluate partial stories until at least 50\% of the planned plot length is reached, preventing premature scoring. This "don’t judge a book by its cover" approach helps avoid discarding branches that might start slowly but improve later. Furthermore, early evaluations could misrepresent the potential of partially developed stories, as seen in Figure \ref{fig:model_accuracy_over_text_portion}. Once a state is past that threshold, the value model $V(s)$ estimates quality based on the partially generated content, providing step-level guidance for MCTS.

\paragraph{State Selection Strategy:} We use a hybrid selection strategy combining the UCB criterion (Equation \ref{eq:ucb}), top percentile average value, and maximum values to select states for expansion---in essence, combining UCB with beam search. Specifically, after standard UCB-based selection, we may also include a small number of top-ranked actions by average $Q$-value or by maximum discovered value, thus mixing exploitation of known high-value expansions with the more balanced exploratory UCB-based picks.

\paragraph{Expansion Rate Adjustment:} We adjust the number of actions per state $\kappa$ dynamically based on the iteration count $k$:

\begin{equation} 
	\kappa = \begin{cases} 
		300 & \text{if } k = 1 \\ 
		8   & \text{if } k = 2 \\ 
		2   & \text{if } k > 2 
	\end{cases}
\end{equation}

\paragraph{Parallel Expansions:} To accelerate search, we do not restrict ourselves to expanding just one node per iteration. Instead, in each MCTS iteration, we select multiple nodes in parallel (e.g., up to 100), expand and simulate them, then perform backpropagation for each. By parallelizing selection and expansion steps, we can rapidly grow the search tree in a single round.

\paragraph{Empirical Tuning:} These hyperparameter choices were informed by initial pilot experiments on 8-level expansions. We did not perform extensive hyperparameter sweeps due to limited compute. The chosen combination of large initial branching, deferred evaluation, and parallel expansions proved sufficient to discover high-value stories within our resource constraints.

\section{ORPO Fine-Tuning: Implementation Details and Dataset Construction}\label{app:orpo_details}

\paragraph{Pairwise Preference Dataset Generation and Preference-Based Fine-Tuning with ORPO}\label{app:pairwise-generation}

\noindent To construct the preference dataset for ORPO, we begin by examining each parent node $s$ in the MCTS-generated search tree $\mathcal{T}$. For each child action $a\in\mathcal{A}(s)$ with successor node $s'$, we retrieve its action-value estimate $Q(s,a)$ (see Equation \ref{eq:QsaUpdate}). We then form ordered pairs $\bigl(\bigl(s,a_i\bigr), \bigl(s,a_j\bigr)\bigr)$ in which $a_i$ is deemed preferable to $a_j$ if the difference in their action values exceeds a minimum threshold, $\Delta Q_{\min}$, i.e., 
$
\Delta Q \;=\; Q(s,a_i)\;-\;Q(s,a_j)\;\ge\;\Delta Q_{\min},
$
and if $Q(s,a_i)$ is above $0.5$, indicating that $a_i$ leads to an overall "good" trajectory. Next, we compute a score for each pair to balance the absolute action-value of $a_i$ with its advantage over $a_j$. Concretely, 
$
\text{Score} \;=\; \beta_{\text{tradeoff}} \cdot Q(s,a_i)\;+\;\bigl(1-\beta_{\text{tradeoff}}\bigr)\;\Delta Q,
$
where $\beta_{\text{tradeoff}}\in [0,1]$ controls the relative weight of these two terms. We rank all pairs by this score and select the top pairs per parent node for our training set. By doing so, we deliberately overrepresent higher-value actions $a_i$, which is crucial because ORPO leverages both a relative odds-ratio term and a supervised fine-tuning term (Negative Log-Likelihood). This overrepresentation ensures that the model is more frequently exposed to desirable expansions.

The ORPO objective combines the supervised fine-tuning (SFT) loss with a relative ratio loss (LOR) based on the log odds ratio between chosen and rejected responses:
\begin{equation}
	\mathcal{L}_{\text{ORPO}} = \mathcal{L}_{\text{SFT}} + \beta \cdot \mathcal{L}_{\text{OR}}
\end{equation}
where $\mathcal{L}_{\text{SFT}}$ is the Negative Log-Likelihood (NLL) loss used for supervised fine-tuning. And $\mathcal{L}_{\text{OR}}$ is the relative ratio loss defined as:
\begin{equation}
	\mathcal{L}_{\text{OR}} = - \log \sigma \left( \log \frac{p_{\theta}(a^+ | s) / (1 - p_{\theta}(a^+ | s))}{p_{\theta}(a^- | s) / (1 - p_{\theta}(a^- | s))} \right)
\end{equation}
Here:
$p_{\theta}(a | s)$ is the probability assigned by the policy model with parameters $\theta$ to action $a$ given state $s$.
$(a^+, a^-)$ are the chosen and rejected actions, respectively.
The hyperparameter $\beta$ balances the contributions of the SFT loss and the relative ratio loss.

In practice, each node $s$ in the search tree may have multiple child actions $\mathcal{A}(s) = \{a_1, a_2, \dots\}$. Consequently, a single "preferred" action $a_1$ may appear in multiple pairwise comparisons, e.g., $\bigl((s, a_1), (s, a_8)\bigr)$, $\bigl((s, a_1), (s, a_7)\bigr)$, $\bigl((s, a_1), (s, a_6)\bigr)$. This repeated presence signals to the policy that $a_1$ consistently outperforms various alternatives from the same parent $s$, amplifying its probability during fine-tuning. Conversely, if two actions (e.g., $(s, a_1)$ and $(s, a_2)$) are found in roughly the same number of "winning" pairs at node $s$, they will tend toward equal learned probabilities (via the SFT part of ORPO). By selectively allowing genuinely higher-value actions $(s, a)$ to appear more often in these pairwise tuples, we ensure that ORPO's supervised (NLL) component places stronger preference weights on the top-performing actions. 

\paragraph{Unsloth + QLoRA Fine-Tuning}\label{app:unsloth-lora}
\begin{figure}[htb]
	\centering
	\includegraphics[width=0.9\columnwidth]{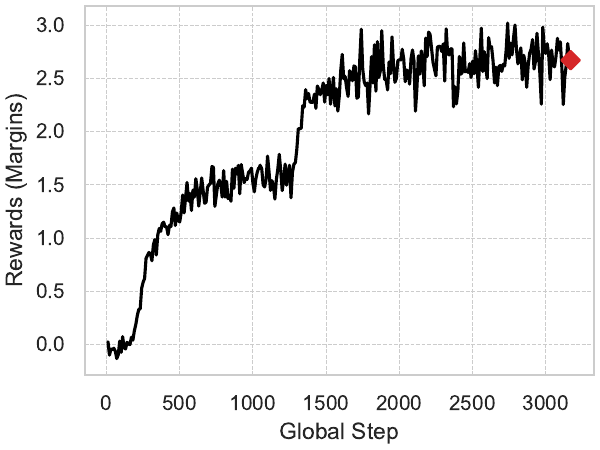}
	\caption{Average training margin against the number of policy updates. Higher margins indicate a stronger separation between chosen (preferred) vs. rejected (disfavored) actions. The final validation-set margin is shown in red.}
	\label{fig:orpo-reward-margin}
\end{figure}
\noindent To enable fine-tuning under our computational constraints, we use Unsloth  \citep{han_unsloth_2023} with 4-bit Quantized Low-Rank Adaptation (QLoRA) \citep{hu_lora_2021}. We set the LoRA rank to $r=$ 64 and scaling factor $\alpha=$ 64, applying LoRA to all linear layers. Fine-tuning via ORPO was conducted over two epochs using $\beta=$ 0.7.

Figure \ref{fig:orpo-reward-margin} illustrates how the reward margin evolves after Round 1 fine-tuning. Initially, it rises from near 0 to $\approx \text{1.5}$ by 1200 updates in a near-sigmoidal trend, then jumps to $\approx \text{2.4}$. It subsequently plateaus around 2.7 at the final iteration. On held-out pairs, the margin reaches 2.67, and the fine-tuned policy model achieves 0.98 accuracy in preferring the higher-value actions.

Finally, the fine-tuned policy model was saved in the Q5\_K\_M quantized format, a mixed-precision scheme that applies a predominantly 5-bit block-wise quantization while allocating 6-bit blocks for particularly sensitive parameters.

\section{Additional Value Model Details}\label{app:value_model_details}

\paragraph{Semantic Coherence Score}
\begin{figure*}
	\centering
	\begin{subfigure}[t]{\textwidth}
		\centering
		\hspace*{-0.04\textwidth}
		\includegraphics[width=1.08\textwidth]{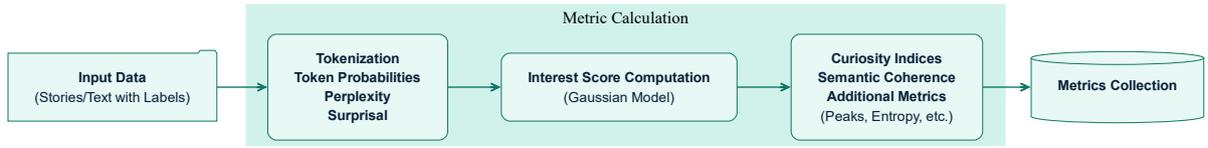}
		\caption{}
	\end{subfigure}
	\begin{subfigure}[t]{\textwidth}
		\centering
		\hspace*{-0.04\textwidth}
		\includegraphics[width=1.08\textwidth]{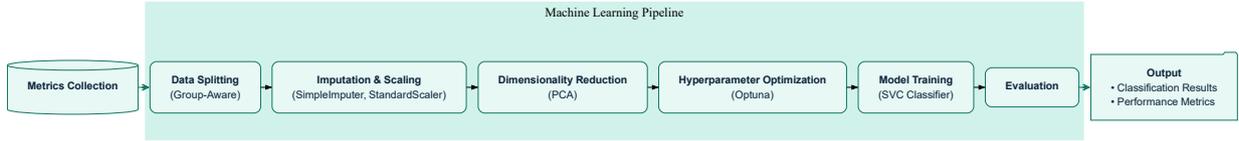}
		\caption{}
	\end{subfigure}
	\caption{(a) Overview of the process to generate measures and metrics from the story text. (b) Overview of the Machine Learning pipeline to produce the value model.}
	\label{fig:stacked_figures_vertical}
\end{figure*}

\noindent To measure thematic consistency across the plot, we encode each sentence into a high-dimensional embedding vector using a pretrained SentenceTransformer model, then compute all pairwise cosine similarities. The semantic coherence score is the average pairwise similarity across all sentences:
\begin{equation} 
	\text{Coherence Score} = \frac{2}{n(n-1)} \sum_{i=1}^{n-1} \sum_{j=i+1}^{n} \frac{E_i \cdot E_j}{|E_i||E_j|}.
\end{equation}
Here, $n$ is the number of sentences, and $E_i$ is the embedding of the $i$th sentence. By considering all sentence pairs, we aim to capture both local and global coherence, such that stories featuring delayed thematic connections are not penalized prematurely.

\paragraph{Additional Features}

\noindent Beyond the curiosity index and coherence score, we derive additional features capturing story dynamics from the surprisal trajectory. These include the frequency and average height of surprisal peaks (local maxima that may indicate plot twists), the variability in intervals between these peaks, and both the average and variance of window-averaged surprisal gradients throughout the plot. In total, we extract 14 features that characterize story engagement, complexity, pacing, and thematic integration.

\paragraph{Full Value Model Training Details}
\begin{figure}[htb]
	\centering
	\includegraphics[width=0.9\columnwidth]{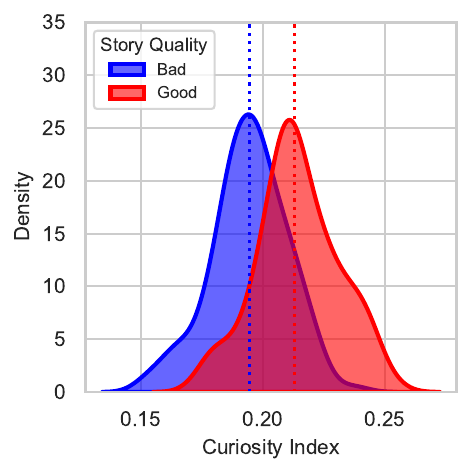}
	\caption{Distribution of the curiosity index for "good" and "bad" stories as encountered by Phi-3.5 3B. The median is the dashed line.
	}
	\label{fig:distribution_curiosity}
\end{figure}
\noindent We train a Support Vector Classifier (SVC) to distinguish between "good" and "bad" stories based on the extracted features. The dataset primarily comprises stories labeled by human judgments. Additionally, we purposefully include a set of low-quality stories generated by smaller LLMs (GPT-3.5 \citep{brown_language_2020}, Gwen 2 7B \citep{yang_qwen2_2024}, Mixtral 8x7B \citep{jiang_mixtral_2024}, and Llama 3 7B \citep{grattafiori_llama_2024}). All stories are first normalized into the same structure: a bullet point list of $K=$ 32 elements outlining the whole plot using GPT-4o. We evaluate these plots at multiple completion levels (50\%, 63\%, 75\%, 88\%, and 100\% of the intended length) to account for evolving story quality.

Our simulation model likewise generates plots in a bullet-point list format that aligns with the value model's training data. By maintaining this alignment, we ensure that the plots generated during MCTS are evaluated consistently with the value model dataset. However, despite this structural consistency, subtle stylistic differences still arise between GPT-4o-generated bullet points and those produced by our simulation model.

To prevent data leakage, we apply group-aware splitting to maintain class distribution across training, validation, and test sets.

We preprocess features via imputation, standardization, and dimensionality reduction using Principal Component Analysis (PCA). We optimize hyperparameters (including regularization parameters $C$, kernel coefficient $\gamma$, number of PCA components, and maximum iterations) using group-aware stratified repeated k-fold cross-validation to maximize the generalization. Figure \ref{fig:stacked_figures_vertical}a and \ref{fig:stacked_figures_vertical}b illustrate the analytical pipeline in two stages.

During hyperparameter optimization, we employ a loss function that prioritizes a low false positive rate and well-calibrated probabilities, while simultaneously minimizing the standard deviation of precision across cross-validation folds.

\paragraph{Correlation between $V(s)$ and $Q(s, a)$}

\begin{figure}[h]
	\centering
	\includegraphics[width=0.7\columnwidth]{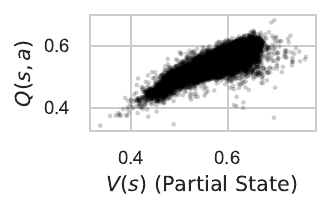}
	\caption{Higher $V(s)$ values tend to correspond to higher $Q(s,a)$, indicating that the value model reliably captures mid-story quality (i.e., step-level value modeling). Each dot (over 52k in total, spanning 18 prompts) is a partial-state expansion that was subsequently explored by MCTS.}
	\label{fig:vs_vs_qsa}
\end{figure}

\noindent In Figure \ref{fig:vs_vs_qsa}, we compare our partial-state value $V(s)$ against the final MCTS returns $Q(s,a)$. The Pearson correlation of $r=0.947$ indicates that higher $V(s)$ strongly aligns with higher $Q(s,a)$, suggesting the learned value function effectively estimates story quality before completion.

\section{Value Model Visualizations}\label{app:results-svc}

\begin{figure}[htb]
	\centering
	\begin{subfigure}[t]{\columnwidth} 
		\centering 
		\includegraphics[width=\columnwidth]{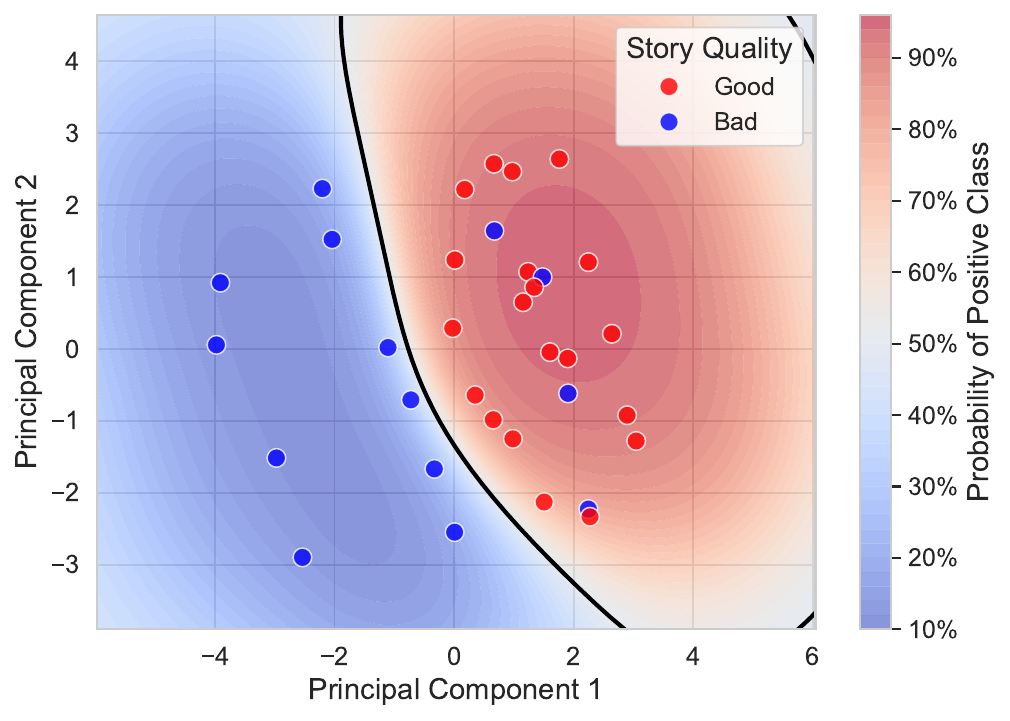}
		\caption{}
	\end{subfigure}
	\begin{subfigure}[t]{\columnwidth}
		\centering
		\includegraphics[width=\columnwidth]{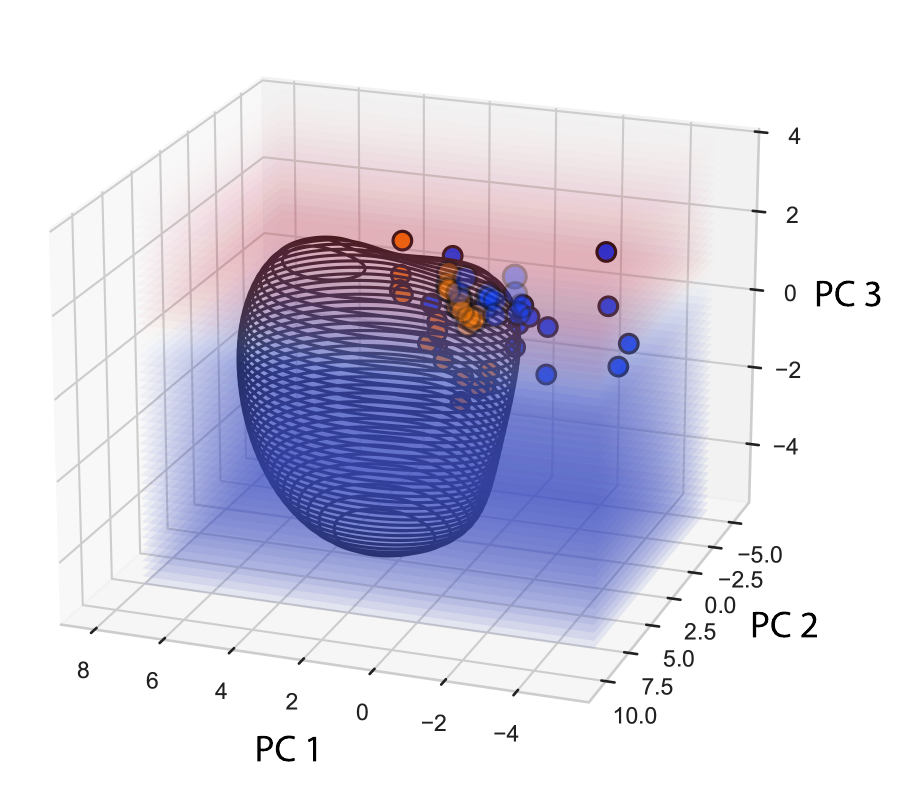}
		\caption{}
	\end{subfigure}
	\caption{(a) The 2D decision boundary (in first two PCA components) at the 0.5 probability contour. (b) A 3D boundary (in first three PCA components) visualized with multiple contour planes. Data points are the test samples, colored by ground-truth labels.}
	\label{fig:svc-vis}
\end{figure}

Figure \ref{fig:svc-vis} shows how the Support Vector Classifier (SVC) separates "good" vs. "bad" plots when projected onto the first two or three PCA components. Panel (a) depicts the 2D boundary at the 0.5 probability contour line, whereas panel (b) provides a 3D volumetric view. Data points (held-out test set) are colored by their true labels.

\begin{figure}[htb]
	\centering
	\includegraphics[width=0.9\columnwidth]{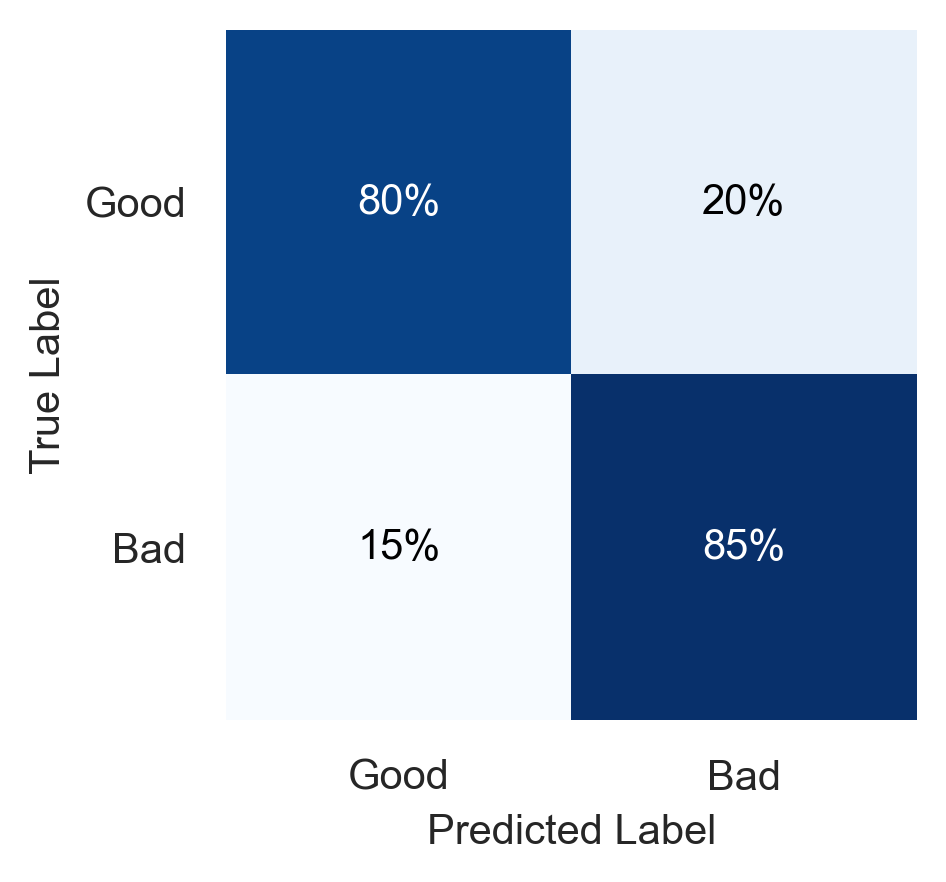}
	\caption{Confusion matrix of the value model's predictions on the held-out test set for plots with $\ge$ 50\% progression.
	}
	\label{fig:confusion_matrix}
\end{figure}

Figure \ref{fig:confusion_matrix} displays the confusion matrix for these predictions, illustrating the model's accuracy in distinguishing "good" from "bad" plots.

\begin{figure}[htb]
	\centering
	\includegraphics[width=0.9\columnwidth]{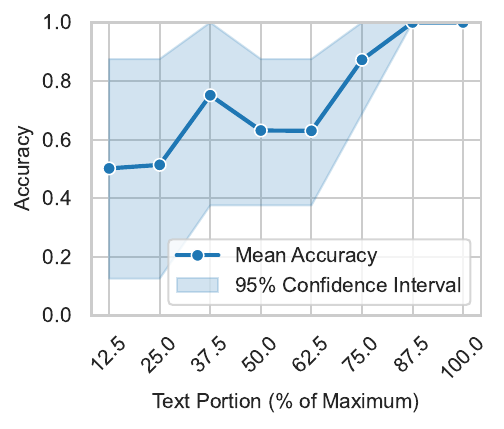}
	\caption{Value model accuracy vs. story completion percentage (95\% CI) on the held-out test set. Accuracy improves with additional context, motivating deferred evaluation}
	\label{fig:model_accuracy_over_text_portion}
\end{figure}

Finally, Figure \ref{fig:model_accuracy_over_text_portion} shows the value model's accuracy as a function of the completion percentage of the story on the held-out test set. The result confirms that the model benefits from more context and motivates our deferred evaluation strategy.


\section{Relative Improvement}\label{app:rel-improvement}

\begin{figure}[htb]
	\centering
	\includegraphics[width=\columnwidth]{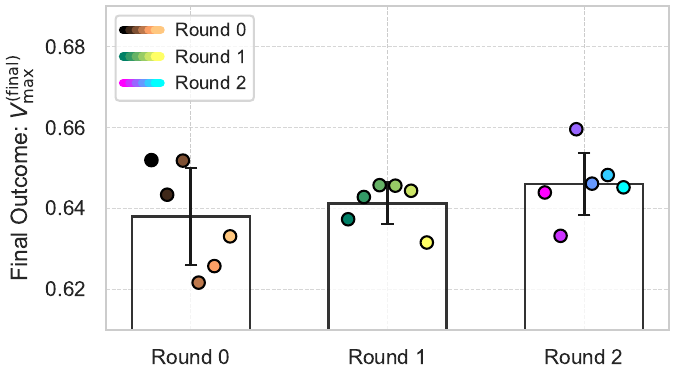}
	\caption{
		Final outcome comparison ($V_{\max}^{(\mathrm{final})}$) across Rounds 0, 1, and 2. Bars show mean, while individual experiments are plotted as distinct dots. The upward trend shows that repeated MCTS+ORPO fine-tuning yields higher-valued stories on average.}
	\label{fig:final-outcome-comparison}
\end{figure}

Figure \ref{fig:final-outcome-comparison} shows the \emph{final} $V_{\max}^{(\mathrm{final})}$ discovered in Round 0, Round 1, Round 2. Each bar represents the average final outcome for that round. Consistent with the faster convergence observed in Section \ref{sec:mcts-orpo-results}, these final outcomes increase from 0.638 $\pm$ 0.012 in Round 0 to 0.641 $\pm$ 0.005 in Round 1 and 0.646 $\pm$ 0.008 in Round 2, indicating a steady upward shift in story quality.

\begin{figure}[htb]
	\centering
	\includegraphics[width=\columnwidth]{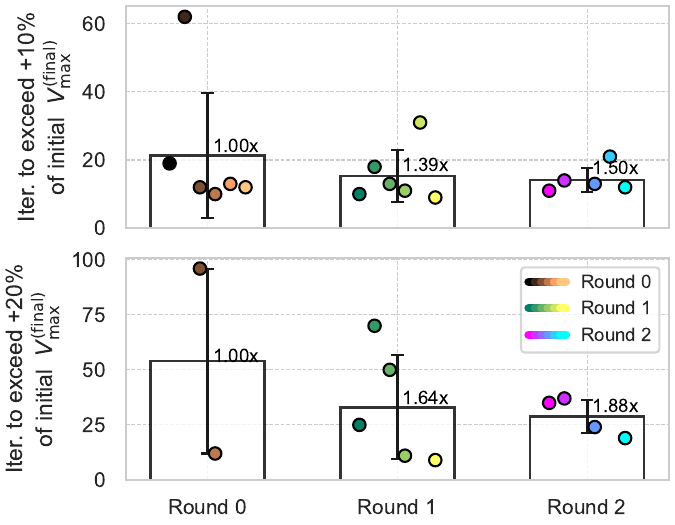}
	\caption{
		We measure how many iterations each round (Round 0, Round 1, Round 2) requires to surpass 
		(top) a 10\% gain and (bottom) a 20\% gain in $V_{\max}^{(\mathrm{final})}$, 
		relative to $k = 8$ (the earliest iteration at which a story is fully generated). Bars indicate the mean number of iterations needed, with error bars for standard deviation, and dots mark individual experiments that reach that threshold. Compared to Round 0, the ORPO-fine-tuned policies in Round 1 and Round 2 reach the 10\% threshold 1.39$\times$ and 1.50$\times$ faster, respectively, and achieve the 20\% threshold 1.64$\times$ and 1.88$\times$ faster.}
	\label{fig:relative-improvement}
\end{figure}
We measured the MCTS iteration count needed in each round (Round 0, Round 1, Round 2) to exceed a specified relative improvement in $V_{\max}^{(\mathrm{final})}$, taking $k=\text{8}$ as our baseline. Figure \ref{fig:relative-improvement} compares the average iteration counts for both 
10\% and 20\% improvement thresholds. Fine-tuning the policy model in Round 1 and Round 2 clearly accelerates the search for high-value stories, significantly reducing the required number of MCTS iterations.

Moreover, the substantial reduction in standard deviations across rounds suggests that fine-tuning yields more consistent performance and faster convergence across diverse prompts, as evidenced by narrower iteration spreads for 10\% and 20\% improvement thresholds.

\section{Evaluation Protocols and A/B Preference Tests}\label{app:ab-comparison}
\begin{table*}
	\centering
	\caption{Participant preferences for each highest- vs. lowest-value story (as labeled by our value model). A "+" indicates the participant favored the model's higher-scored plots, whereas "-" means they chose the lower-scored one. Empty cells denote no response.}
	\begin{tabular}{lcccccccc}
		\toprule
		\textbf{Participant} & \textbf{1} & \textbf{2} & \textbf{3} & \textbf{4} & \textbf{5} & \textbf{6} & \textbf{7} & \textbf{8} \\
		\midrule
		\textbf{Exp 1} & + & + & + & + & - & + & + & - \\
		\textbf{Exp 2} & + & - & + & + & + & + & + & - \\
		\textbf{Exp 3} & - & - & + & + &   & + & + & - \\
		\textbf{Exp 4} & + & + & + & - &   & + & - & - \\
		\bottomrule
	\end{tabular}
	\label{tab:a-b-human-test}
\end{table*}
\paragraph{A/B Preference Test with Human Participants}

\noindent We recruited eight university-educated participants (mean age around 30 years) from the authors' social circles. All were fluent in English, were informed about the study's nature, and gave consent. Prior to viewing any stories, each participant received a short "Participant Manual" that explained the study's purpose and how to interpret the bullet-point story outlines. The manual emphasized that the stories came from a relatively simple language model, that they should evaluate the plots rather than the writing style, and that their responses would be used only for research evaluation and not for further model training.

Following these instructions, participants were presented with pairs of story plot outlines, each consisting of 32 bullet points. They read both outlines in each pair and indicated which plot they preferred overall. The stories were labeled "A" or "B" in random order, and neither the participants nor did the author knew which had been scored higher or lower by the value model. Participants' choices were kept confidential and are not visible to others. Participants did not see each other's responses, nor did they have access to any identifying information about others who took part. They could also (optionally) provide free-text comments explaining their selections.

\paragraph{Experimental Design}

\noindent We selected the trajectories that our system had rated highest and lowest in plot quality (i.e., "best" and "worst" under our value model) for each pair. MCTS pruning often discards extremely poor story paths early, so "lowest-value" refers to the least promising among those that reached full completion; it does not necessarily mean the story is objectively bad. Across four such pairs, an average of eight participants responded per pair, yielding 30 total pairwise ratings.

For confidentiality, we assigned numerical labels (1--8) to participants in the table below that do not reflect the alphabetical ordering used in the Acknowledgments.

\paragraph{Results and Analysis}

\noindent Table \ref{tab:a-b-human-test} reports participants' preferences. A "+" entry indicates that the participant favored the story the model rated higher, wheras a "-" indicates disagreement with the value model's ranking. Excluding two empty entries left 30 valid responses, 20 of which ("+") agreed with the model (66.7\%). Since the sample size was relatively small, we did not conduct a formal inference test (e.g., a chi-square test) to measure significance. Notably, one participant consistently selected the lowest-rated expansions. Removing that individual (outlier) yields 26 valid responses, 20 of which ("+") favored the model's top-ranked story, raising the overall preference to 76.9\%. Although these stories may differ only subtly in quality (since low-value candidates are not further expanded), the trend indicates that our value model's "best" trajectory often aligns with human preference.

\paragraph{Limitations}

\noindent Our participant pool was recruited via convenience sampling from the authors' acquaintances, and it was relatively homogeneous in age and educational background. This lack of diversity may limit the generalizability of our results. Although the story labels were randomized to mitigate expectancy effects, participants were aware that the texts were generated by an LLM, which could have influenced perceptions. Furthermore, some participants may hold negative (or positive) predispositions toward AI, potentially creating biases in their choices. For instance, we observed that one participant consistently disagreed with the model's rankings, which might reflect an "anti-AI" sentiment or a consistent personal preference. Because our study size was modest and participants' attitudes toward AI were not formally measured, we cannot rule out such effects. Finally, as in any convenience-sample study, it is possible that unstated biases or other external factors influenced the results in ways that we have not accounted for.
\vfill

\pagebreak
\paragraph{GPT‑4o Absolute Rating Instruction}

\noindent We also asked GPT‑4o to provide absolute rating:
\begin{textbox}\label{app:textbox:gpt4o-eval}
	\begin{small}
	Please evaluate the following story:\\
	
	\{Story Text\}\\
	\\
	Evaluation Criteria:\\
	
	1. Plot Structure (1-10):
	- Does the story have a clear beginning, middle, and end?
	- Is the plot coherent and logically structured?\\
	
	2. Tension (1-10):
	- Does the story build suspense or keep the reader engaged?
	- Are there conflicts or obstacles that the characters must overcome?\\
	
	3. Originality (1-10):
	- Is the story unique or does it offer a fresh perspective?
	- Does it avoid clichés and predictable plot lines?\\
	
	4. Character Development (1-10):
	- Are the characters well-developed and believable?
	- Do they undergo significant growth or change?\\
	
	5. Overall Impact (1-10):
	- Does the story leave a lasting impression?
	- How effective is it in conveying emotions or themes?\\
	
	6. Theme (1-10):
	- Is there a clear central theme or message?
	- How well is this theme integrated into the story?\\
	
	7. Conflict (1-10):
	- Is the central conflict compelling and well-executed?
	- Does it drive the story forward?\\
	
	8. Pacing (1-10):
	- Does the story flow smoothly?
	- Are there any slow or rushed sections?\\
	
	9. Style and Voice (1-10):
	- Is the writing style appropriate for the story?
	- Does the author have a distinctive voice?\\
	
	Please provide a brief justification for each rating and then assign a score from 1-10 for each category.\\
	
	Finally, after you have gone through all 9 points and gave your text-based justification, please generate a JSON:\\
	
	**JSON Output**: Please provide a structured output in JSON format with the following keys, where each value must be an integer between 1 and 10:\\
	\\
	json\\
	\{\\
	"Plot Structure": {{ (1-10) }},\\
	"Tension": {{ (1-10) }},\\
	"Originality": {{ (1-10) }},\\
	"Character Development": {{ (1-10) }},\\
	"Overall Impact": {{ (1-10) }},\\
	"Theme": {{ (1-10) }},\\
	"Conflict": {{ (1-10) }},\\
	"Pacing": {{ (1-10) }},\\
	"Style and Voice": {{ (1-10) }}\\
	\}
\end{small}
\end{textbox}

\paragraph{A/B Preference Test with o1}\label{app:o1-comparison-results}

\noindent We used o1 for a pairwise plot comparison between COS(M+O)S and the Llama 3.2 3B baseline (no MCTS).
\begin{textbox}\label{app:textbox:o11}
	\begin{small}
	\textbf{User}:
	Which story do you prefer? Do NOT judge the bullet-point structure, choose purely by plot (and not by writing style or format).\\

	"\\
	As I stood at the bow of the small sailboat, the salty spray of the ocean mingling with the scent of seaweed, I felt the gentle caress of a breeze brush against my skin. It was as if the wind itself was whispering secrets in my ear, carrying the faintest hint of a message on its breath. I turned to see what had caught my attention, and that's when I saw it - a crumpled piece of paper, torn from a notebook and lifted by the wind into the air.\\
	
	I watched, mesmerized, as the paper danced on the breeze, its edges fluttering like a bird's wings. The wind seemed to be carrying it towards me, and I reached out a hand to catch it. As my fingers closed around the paper, I felt a shiver run down my spine. The paper was damp and worn, with a faint scent of ink and parchment.\\
	
	I unfolded the paper, and a message scrawled in hasty handwriting leapt out at me:\\
	
	"Meet me at the old lighthouse at midnight. Come alone."\\
	
	The words sent a thrill through me, but also a sense of trepidation. Who could have written this? And why? I looked around, but the sailboat was deserted, the only sound the creaking of the wooden hull and the distant cry of gulls.\\
	
	As I stood there, trying to make sense of the message, the wind picked up, and the sailboat began to heel to one side. I stumbled, catching myself on the rail, and that's when I realized that the wind was not just blowing the paper - it was also carrying me away.\\
	
	The sailboat surged forward, its sails filling with wind as we picked up speed. I felt a rush of exhilaration mixed with fear as the world around me blurred into a green and blue haze. The lighthouse loomed ahead, its towering spire reaching for the stars like a skeletal finger.\\
	
	I clung to the rail, my heart pounding in my chest, as the sailboat bore me towards the lighthouse. The wind howled around me, whipping my hair into a frenzy, and I felt myself being swept away by its power.\\
	
	As we approached the shore, I saw a figure waiting for me on the beach - a woman with long, flowing hair and eyes that seemed to see right through me. She beckoned me towards her, and I felt a strange sense of recognition, as if I had known her all my life.\\
	
	"Who are you?" I asked, my voice barely above a whisper.\\
	
	"I am the one who wrote the message," she replied, her voice low and husky." And you, my dear, are the one who has been swept away by the wind."\\
\end{small}
\end{textbox}
\begin{textbox}\label{app:textbox:o12}
\begin{small}
	As I looked into her eyes, I felt a shiver run down my spine. I knew that I was in the midst of something much bigger than myself, something that would change me forever. And I knew that I was ready to follow the wind, wherever it might lead.\\
	"\\
	\\
	"\\
	• The sailboat, "Maverick's Revenge," cuts through the calm morning waters of the Caribbean, its white sails billowing in the gentle breeze. \\
	• Captain James "Hawk" Hawkins stands at the helm, his eyes fixed on the horizon, his weathered face a map of lines and creases. \\
	• A faint rustling sound comes from the cargo hold, where a small, crumpled piece of paper has been tossed aside, its contents unknown. \\
	• Unbeknownst to Hawk, the paper was left behind by a mysterious stranger, who had vanished into the night, leaving behind only the whispered words: "Beware the wind's dark side."\\
	• As Hawk scans the horizon, his first mate, Tom, approaches him with a concerned expression, his eyes darting towards the cargo hold. \\
	• "Cap'n, I think we've got a problem," Tom says, his voice low and urgent, "that stranger who was on board last night, he's gone missin'." \\
	• Hawk's expression turns skeptical, but Tom's words are reinforced by the sound of footsteps coming from below deck, heavy and deliberate. \\
	• The footsteps stop at the cargo hold, and Hawk's eyes narrow as he calls out, "What's goin' on down there, Tom?"\\
	• The sun beats down on the deck, but Hawk's attention is drawn to the sky, where a dark shape seems to be gathering, its edges indistinct but ominous. \\
	• Tom follows his gaze, his brow furrowed in concern, and together they watch as the shape grows larger, its presence seeming to fill the air. \\
	• The wind picks up, carrying the faint scent of salt and ozone, and Hawk's instincts tell him that something is about to go terribly wrong. \\
	• As the shape looms closer, Hawk's eyes lock onto the horizon, where a massive storm cloud is taking shape, its leading edge a deep, foreboding grey.\\
	• As the storm cloud draws near, Hawk's gaze falls upon Tom, who stands frozen at the rail, his eyes fixed on the horizon with a mixture of fear and fascination. \\
	• The air is heavy with tension as Hawk notices that Tom's eyes seem to be drawn to something on the deck, something that Hawk can't see.\\
\end{small}
\end{textbox}
\begin{textbox}\label{app:textbox:o13}
\begin{small}
	• Hawk's instincts tell him that Tom is being manipulated, but by whom, and what, he can't quite grasp. \\
	• The wind howls louder, and the storm cloud's leading edge breaks through the surface, casting a dark, churning shadow over the deck, where Tom's eyes remain fixed on something that seems to be waiting for him.\\
	• The storm rages on, its fury unrelenting, as Tom's eyes lock onto the cargo hold, his face twisted in a mixture of terror and fascination. \\
	• The voice in his head is a low, raspy whisper, speaking a single word: "Mine." \\
	• The words are followed by a vision of the mysterious stranger, his eyes black as coal, his skin stretched taut over his skull. \\
	• The vision fades, and Tom's eyes snap back into focus, but the voice remains, its presence echoing through his mind like a dark, relentless drumbeat.\\
	• The stranger's face begins to shift and writhe, his features melting like wax in a furnace, revealing a twisted, inhuman visage beneath. \\
	• Tom's eyes go wide as he realizes the truth: the stranger is a vessel, a puppet for an ancient, malevolent force that has been awakened by the storm. \\
	• The force, known only as "The Devourer," begins to stir, its presence coursing through the ship's hull like a dark, living thing. \\
	• Tom's eyes widen in horror as he remembers the night he met the stranger, the night he made a pact with a dark and malevolent force. \\
	• The memory is like a key turning in a lock, unlocking a floodgate of terror and regret, as Tom realizes he's been drawn back into a world he thought he'd left behind. \\
	• The storm cloud seems to be closing in, its darkness coalescing into a presence that's both familiar and yet utterly alien. \\
	• Tom's mind reels as he whispers a single word: "No," the sound lost in the howling wind, as the stranger's eyes flash into his mind, burning with an otherworldly intensity.\\
	• Tom's face goes white as he staggers backwards, his eyes fixed on the cargo hold as if he's waiting for something to emerge. \\
	• His hand reaches out, as if to grasp something, but instead he clutches at his own wrist, his eyes searching for a mark or a scar that's not there. \\
	• The storm rages on, but Tom's attention is focused inward, his mind reeling with memories and regrets. \\
	• His grip on reality begins to slip, and he whispers a single word: "Run."\\
	• The storm cloud bursts open, revealing a figure standing at its center, its eyes blazing with an otherworldly energy.\\
\end{small}
\end{textbox}

\begin{textbox}\label{app:textbox:o14}
	\begin{small}
		• Tom recognizes the stranger, and with a surge of adrenaline, he takes a step forward, his eyes locked on the figure's.\\		
		• The stranger's gaze holds Tom in place as it speaks a single word: "Welcome," and the darkness that has haunted Tom for so long begins to dissipate. \\
		• As the storm dissipates, Tom stands tall, his eyes shining with a newfound understanding, and he whispers a single word: "Free."\\
		"\\
	\end{small}
\end{textbox}
\begin{textbox}\label{app:textbox:o15}
	\begin{small}
	\textbf{o1}:\\
	Purely on the basis of plot (ignoring writing style or format), I’d pick the \textbf{second version}. It has a larger sense of conflict—there’s a lurking supernatural threat, a secret pact, and higher stakes aboard the sailboat. The ominous storm and the mystery of the stranger give it an intense, unfolding tension that grabbed my attention a bit more than the simpler, though still intriguing, mystery in the first story.
\end{small}
\end{textbox}
\begin{textbox}\label{app:textbox:o16}
	\begin{small}
	\textbf{User}:\\
	Give both stories a rating out of 100. Be realistic and truthful. You act as a harsh critic\\
\end{small}
\end{textbox}
\begin{textbox}\label{app:textbox:o17}
\begin{small}
\textbf{o1}:\\
\textbf{Story 1 Rating: 65/100}

- Reasoning (Plot-Focused): While it sets a mysterious tone, the plot is quite straightforward and resolves too quickly. There’s an intriguing hook (the cryptic note and the mysterious woman), but the conflict never truly escalates before it’s essentially resolved. Overall, it feels a bit underdeveloped, as if it were just the first chapter of a longer mystery rather than a fully fleshed-out storyline.

\textbf{Story 2 Rating: 80/100}
 
- Reasoning (Plot-Focused): This plot has a stronger sense of tension and a deeper conflict, with supernatural elements and a dark backstory. The looming storm and the pact made by Tom introduce higher stakes, which create a more compelling scenario. However, it relies on some familiar tropes (the cursed pact, the ominous stranger), and the resolution wraps up quickly. It’s more engaging than Story 1 but could still benefit from additional depth and refinement.
\end{small}
\end{textbox}

\paragraph{Extracted Scores Across Baseline (No MCTS) vs. COS(M+O)S}

\noindent Table \ref{tab:o1-comparison-results-llama3b} shows the baseline/MCTS scores for all rounds (R0--R2) and experiments (Exp 1--6) in the pairwise o1 comparison against Llama 3.2 3B and in Table \ref{tab:o1-comparison-results-llama70b} against Llama 3.1 70B (using the Llama 3.2 3B backend). Table \ref{tab:o1-comparison-results2-llama3b} lists the difference (MCTS minus baseline) per experiment, illustrating that while a few cases slightly favored the baseline, most showed sizable MCTS gains (up to +20).
\begin{table}[htb]
	\centering
	\caption{Side-by-side o1-based ratings (0--100) for Llama 3.2 \textbf{3B} baseline (no MCTS) vs. COS(M+O)S across Rounds 0--2 and experiments (Exp 1--6). Each cell shows baseline/COS(M+O)S.}
	\begin{tabular}{lcccccc}
		\toprule
		& Exp 1 & Exp 2 & Exp 3 & Exp 4 & Exp 5 & Exp 6 \\
		\midrule
		\textbf{R0} & 72/82 & 70/82 & 78/92 & 81/73 & 65/80 & 70/80 \\
		\textbf{R1} & 75/82 & 75/84 & 68/80 & 75/82 & 70/80 & 70/79 \\
		\textbf{R2} & 70/84 & 82/72 & 78/70 & 65/75 & 65/85 & 65/80 \\
		\bottomrule
	\end{tabular}
	\label{tab:o1-comparison-results-llama3b}
\end{table}
\begin{table}[!ht]
	\centering
	\caption{Side-by-side o1-based ratings (0--100) for Llama 3.1 \textbf{70B} baseline (no MCTS) vs. COS(M+O)S across Rounds 0--2 and experiments (Exp 1--6). Each cell shows baseline/COS(M+O)S.}
	\begin{tabular}{lcccccc}
		\toprule
		& Exp 1 & Exp 2 & Exp 3 & Exp 4 & Exp 5 & Exp 6 \\
		\midrule
		\textbf{R0} & 65/50 & 78/72 & 55/75 & 65/75 & 60/75 & 65/75 \\
		\textbf{R1} & 72/60 & 65/75 & 75/55 & 65/75 & 78/65 & 68/82 \\
		\textbf{R2} & 65/60 & 75/60 & 65/80 & 78/55 & 60/70 & 60/80 \\
		\bottomrule
	\end{tabular}
	\label{tab:o1-comparison-results-llama70b}
\end{table}
\begin{table}[!ht]
	\centering
	\caption{Difference table (COS(M+O)S minus baseline). Positive values indicate where COS(M+O)S outscored the Llama 3.2 \textbf{3B} baseline.}
	\begin{tabular}{lcccccc}
		\toprule
		& Exp 1 & Exp 2 & Exp 3 & Exp 4 & Exp 5 & Exp 6\\
		\midrule
		\textbf{R0} & +10 & +12 & +14 & -8 & +15 & +10 \\
		\textbf{R1} & +7 & +9   & +12 & +7 & +10 & +9 \\
		\textbf{R2} & +14 & -10 & -8 & +10 & +20 & +15\\
		\bottomrule
	\end{tabular}
	\label{tab:o1-comparison-results2-llama3b}
\end{table}

\end{document}